\documentclass[10pt,twocolumn,letterpaper]{article}

\usepackage{iccv}
\usepackage{times}
\usepackage{epsfig}
\usepackage{graphicx}
\usepackage{amsmath}
\usepackage{amssymb}
\usepackage{multirow}
\usepackage{wrapfig}

\usepackage[dvipsnames]{xcolor}
\colorlet{LightRubineRed}{RubineRed!70!}
\colorlet{Mycolor1}{green!10!orange!90!}
\definecolor{Mycolor2}{HTML}{00F9DE}

\usepackage[breaklinks=true,bookmarks=false]{hyperref}

\iccvfinalcopy 

\def\iccvPaperID{****} 

\ificcvfinal\fi

\begin{document}

\title{Solving Inefficiency of Self-supervised Representation Learning}

\author{Guangrun Wang$^{1,2}$ Keze Wang$^4$ \quad Guangcong Wang$^3$ \quad Philip H.S. Torr$^2$ \quad Liang Lin$^1$\thanks{Corresponding Author.}  \\
\small$^1$ Sun Yat-sen University \quad
\small$^2$ University of Oxford \quad
\small$^3$ Nanyang Technological University \quad
\small$^4$DarkMatter AI Research \\{\tt\small \{wanggrun,wanggc3,kezewang\}@gmail.com,} \quad {\tt\small philip.torr@eng.ox.ac.uk, linliang@ieee.org}
}


\maketitle
\ificcvfinal\thispagestyle{empty}\fi

\begin{abstract}
Self-supervised learning (especially contrastive learning) has attracted great interest due to its huge potential in learning discriminative representations in an unsupervised manner. Despite the acknowledged successes, existing contrastive learning methods suffer from very low learning efficiency, e.g., taking about ten times more training epochs than supervised learning for comparable recognition accuracy. In this paper, we reveal two contradictory phenomena in contrastive learning that we call {\bf under-clustering} and {\bf over-clustering} problems, which are major obstacles to learning efficiency. Under-clustering means that the model cannot efficiently learn to discover the dissimilarity between inter-class samples when the negative sample pairs for contrastive learning are insufficient to differentiate all the actual object classes. Over-clustering implies that the model cannot efficiently learn features from excessive negative sample pairs, forcing the model to over-cluster samples of the same actual classes into different clusters. To simultaneously overcome these two problems, we propose a novel self-supervised learning framework using a truncated triplet loss. Precisely, we employ a triplet loss tending to maximize the relative distance between the positive pair and negative pairs to address the under-clustering problem; and we construct the negative pair by selecting a negative sample deputy from all negative samples to avoid the over-clustering problem, guaranteed by the Bernoulli Distribution model. We extensively evaluate our framework in several large-scale benchmarks (e.g., ImageNet, SYSU-30k, and COCO). The results demonstrate our model's superiority (e.g., the learning efficiency) over the latest state-of-the-art methods by a clear margin. See Codes\footnote{\textcolor{red}{\url{https://github.com/wanggrun/triplet}}}.
\end{abstract}

\section{Introduction}

\begin{figure}
\centering
\includegraphics[width=0.4\textwidth]{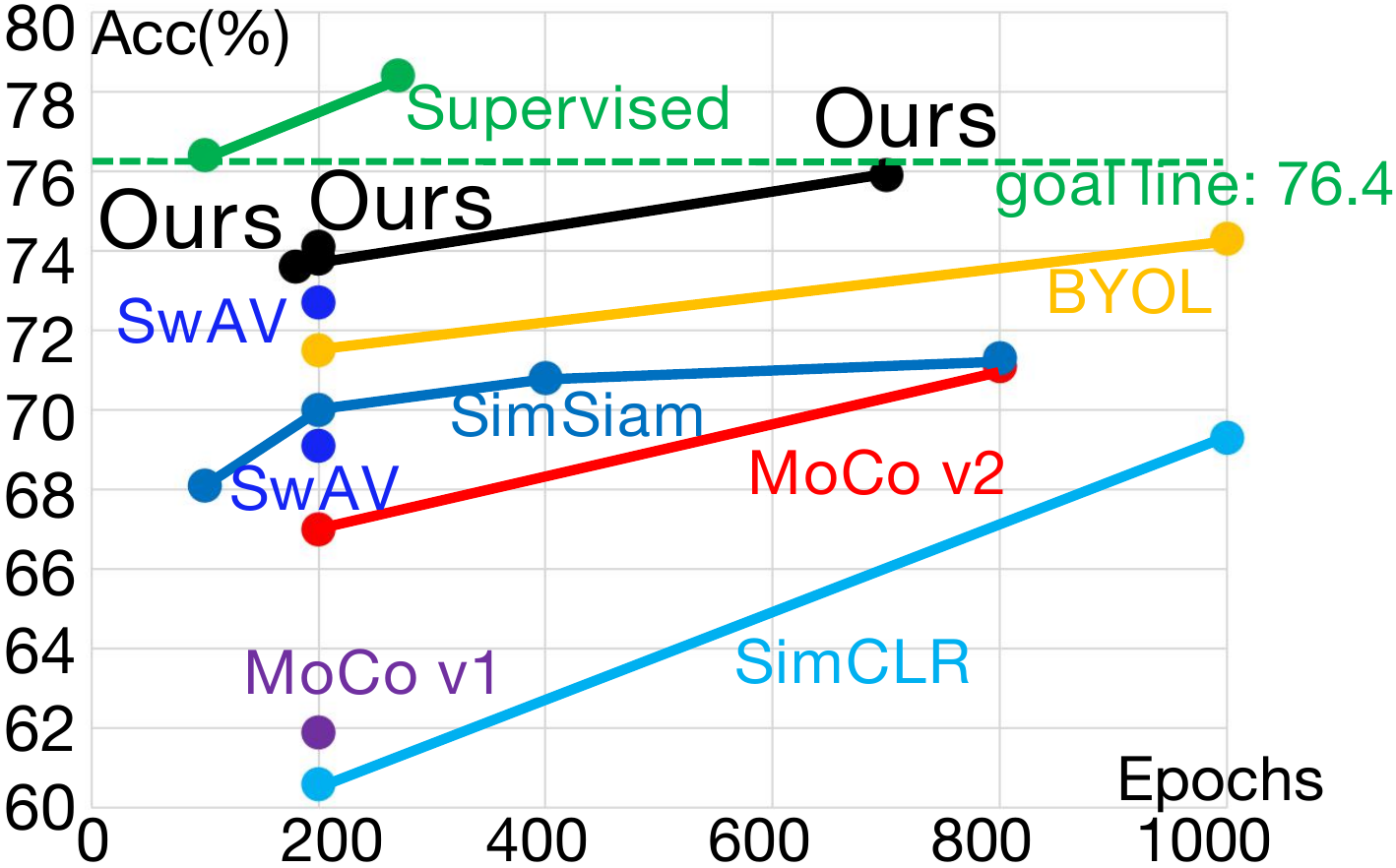}
\caption{A comparison of learning efficiency among different SSL methods using ResNet-50. Here, the x-axis represents the training epochs of SSL, and the y-axis stands for the top-1 accuracy of ImageNet linear evaluation. All methods have lower learning efficiency than supervised learning, but our approach has a significantly higher learning efficiency than the existing SSL methods. (\textbf{best view in color})}\label{fig:intro}
\vspace{-15pt}
\end{figure}

Recently, self-supervised learning (SSL) has shown remarkable results in representation learning. Among them, the results of contrastive learning are most promising in the computer vision tasks. Notable works include MoCo v1/v2 \cite{He2020Momentum_cvpr,Chen2020Improved_arxiv}, SimCLR \cite{Chen2020Simple_arxiv}, BYOL \cite{Grill2020Bootstrap_arxiv}, and SimSiam \cite{Chen2021Exploring_cvpr}. For example, on ImageNet \cite{russakovsky2015imagenet}, the top-1 accuracy of BYOL is 74.3\%, which is close to that of supervised learning, i.e., 76.4\% \cite{wu2016tensorpack,Abadi2015TensorFlow,Jia2014Caffe_acmmm,Paszke2019Pytorch_neurips} (see \emph{``\textcolor{green}{goal line}''} in Figure \ref{fig:intro}). Despite the promising accuracies and high expectations, the learning efficiency of the state-of-the-art SSL methods is about ten times lower than the supervised learning methods. For instance, the supervised learning method typically takes about 100 epochs to train a ResNet50 on ImageNet. In comparison, SimCLR and BYOL have to cost 1,000 epochs, and MoCo v2 needs to cost 800 epochs (See Figure \ref{fig:intro}).

Attempting to address this issue, we rethink existing SSL methods' mechanism and attribute their inherited drawback to two opposing problems, i.e., \textbf{under-clustering} and \textbf{over-clustering}. Specifically, during batch training, contrastive learning randomly crops each image two times to obtain two views and study the similarity between these two views (called a \textbf{positive sample pair} \footnote{e.g., View A and View B of Image X}). Meanwhile, some methods also study the dissimilarity between cross-image views (called \textbf{negative sample pairs}\footnote{e.g., View A of Image X and View B of Image Y}). The optimization objective is to reduce the distance between positive sample pairs and enlarge the distance between negative sample pairs. As suggested by metric learning \cite{Ding2015Deep_pr}, sufficient negative sample pairs are required to guarantee the learning efficiency. Otherwise, lacking negative samples  -- whether due to the GPU memory constraints like SimCLR or (ii) algorithm design like BYOL and SimSiam \cite{Chen2021Exploring_cvpr} -- can make different object categories having overlaps. This is identified as the \emph{under-clustering} problem. One evidence of under-clustering is shown in Table \ref{tab:intro}\footnote{We calculate the class center for each category and compute the distances for every two class centers. These center-to-center distances are averaged to form a class divergence. We keep the variance equal so we can just compare class divergence. The small class divergence in Table \ref{tab:intro} indicates BYOL does suffer under-clustering.}. As a result of under-clustering, SimCLR and BYOL have low learning efficiency because the model cannot efficiently discover the dissimilarity between inter-class samples. On the contrary, excessive negative samples can lead to an opposite problem, i.e., \emph{over-clustering}, which implies the negative samples are false negative and the model over-clusters samples of the same actual categories into different clusters. In an extreme case, there would be 1.28M clusters for ImageNet! One evidence of over-clustering is in Table \ref{tab:intro}\footnote{We use $\Pr(\omega| \mathcal{A})$ (defined in Section \ref{sec:ablation}) to denote the possibility of containing a false negative samples in a batch. The high probabilities in Table \ref{tab:intro} verify MoCo v2 indeed suffers over-clustering.}. Over-clustering also results in low learning efficiency since it encourages dissimilarity between intra-class samples in vain. As reported by \cite{Zhang2017Understanding_iclr,bresler2020corrective}, over-clustering can lead to unnecessary harmful representation learning. For example, \cite{Dosovitskiy2014Discriminativ_neurips,Dosovitskiy2016Discriminative_pami} obtains an unsatisfied performance due to directly clarifying CIFAR-10 into 50K clusters. MoCo v1/v2 cannot further increase the accuracy, even leveraging the momentum to store plenty of negative samples. In summary, existing contrastive learning cannot avoid the under-clustering or over-clustering problems, so their learning efficiency is still low.

To tackle the above under-clustering and over-clustering problems, a few pioneering works have been proposed to analyze the negative samples' role in the contrastive loss \cite{Chuang2020Debiased_neurips,Robinson2021Contrastive_iclr,Cai2020Are_arxiv,Huynh2020Boosting_arxiv}. As opposed to these methods that use over-complicated contrastive losses, we propose an SSL framework using a quite simple truncated triplet loss. Specifically, a triplet loss can maximize the relative distance between the positive and negative pairs for each triplet unit. Having plenty of triplets, we can address the under-clustering problem because wealthy triplets contain rich negative pairs that guarantee a considerable distance between negative sample pairs. Triplet loss largely addresses the under-clustering issue but raises the over-clustering problem. Hence, we propose a novel truncated triplet loss to avoid over-clustering samples from the same category into different clusters. The truncated triplets are ensured with confidence guaranteed by the Bernoulli Distribution model. This significantly improves SSL's learning efficiency and leads to state-of-the-art performance (See Figure \ref{fig:intro}).

In summary, our contribution is three-fold. \begin{itemize}
\item We analyze the existing best-performing contrastive learning methods and attribute their low learning inefficiency to the under-clustering and over-clustering, which result in unnecessary harmful representation learning just to memorize the data.

\item To address the under-clustering and over-clustering problem, we propose a novel SSL framework using a truncated triplet loss. Precisely, we employ a triplet loss containing rich negative samples to address the under-clustering problem, and our triplet loss uses truncated/trimmed triplets to avoid over-clustering, guaranteed by the Bernoulli Distribution model. 

\item Our method significantly improves SSL's learning efficiency and thus leads to state-of-the-art performance in several large-scale benchmarks (e.g., ImageNet \cite{russakovsky2015imagenet}, SYSU-30k \cite{Wang2020Weakly_tnnls}, and COCO 2017 \cite{Lin2014Microsoft_eccv}) and varieties of downstream tasks.

\end{itemize} 

\begin{table}
    \centering
    \small
    \begin{tabular}{c|c|c}
    \hline
   \multirow{2}{*}{Under-clustering (Divergence)} & BYOL & \textbf{\textcolor{Mycolor1}{ Ours}}\\\cline{2-3}
     & 5.3711  & \textbf{\textcolor{Mycolor1}{7.6803}}\\\hline
   \multirow{2}{*}{Over-clustering ($\Pr(\omega| \mathcal{A})$)} &  MoCo v2 & \textbf{\textcolor{Mycolor1}{Ours}}  \\
    \cline{2-3}
       &  1.0 & \textbf{\textcolor{Mycolor1}{0.0110}}  \\\hline
    \end{tabular}
    \caption{\small{Qualitative analysis of over-/ under- clustering. We use $\Pr(\omega| \mathcal{A})$ (the larger, the higher over-clustering risk) and class divergence (the smaller, the higher under-clustering risk) to measure the over-/ under- clustering level, respectively.}}\label{tab:intro}
    \vspace{-15pt}
\end{table}

\begin{figure*}
\centering
\includegraphics[width=0.9\textwidth]{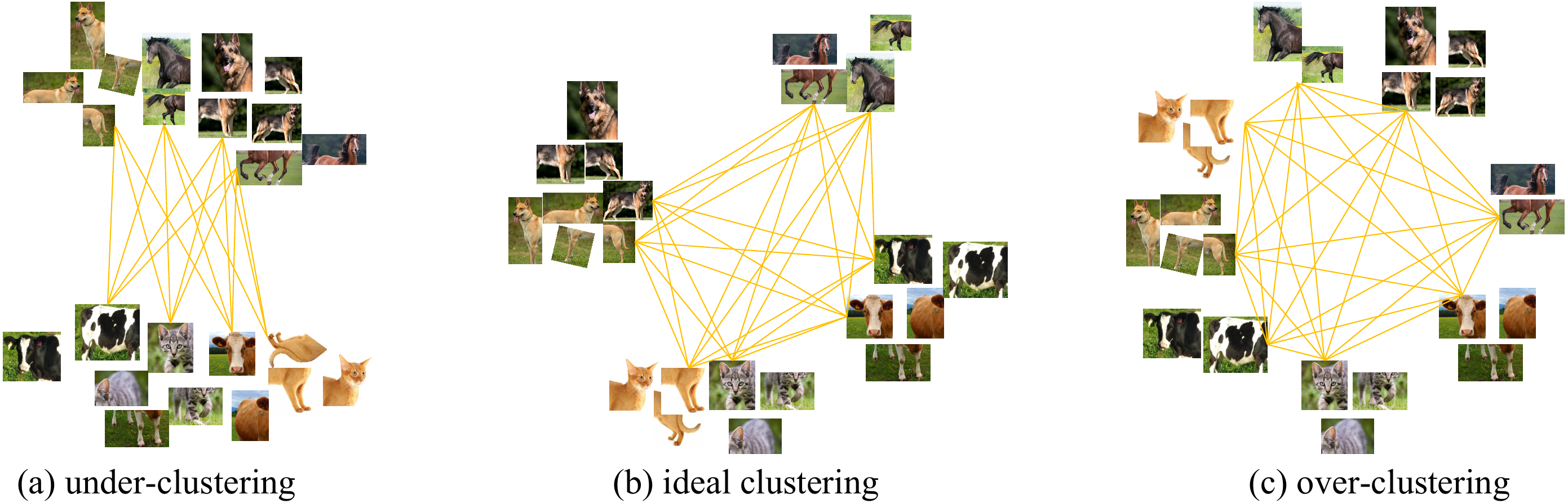}
\caption{Illustration of under-clustering and over-clustering. Each sample pair connected by a yellow line represents a negative pair.}\label{fig:problem}
\vspace{-14pt}
\end{figure*}

\vspace{-11pt}
\section{Related Work}

\textbf{Vanilla SSL.} The recent renaissance of SSL originated from straightforward pretext tasks. Typical pretext tasks included image denoising \cite{Vincent2008Extracting_icml}, image inpainting \cite{Pathak2016Context_cvpr}, patch ordering \cite{Doersch2015Unsupervised_iccv}, solving jigsaw puzzles \cite{Noroozi2016Unsupervised_eccv}, color jittering \cite{Zhang2016Colorful_eccv}, and rotation prediction \cite{Gidaris2018Unsupervised_iclr}. Although these methods contributed to the renaissance of SSL, their learned representations did not generalize well.

\textbf{Contrastive learning.} Currently, the most effective SSL method in computer vision is contrastive learning \cite{Chen2020Simple_arxiv,He2020Momentum_cvpr,Chen2020Improved_arxiv,Grill2020Bootstrap_arxiv,Caron2020Unsupervised_arxiv}, in which the intra-class distances are encouraged to be small, and the inter-class distances are forced to be large\footnote{although the class labels are unknown.}. Plenty of positive and negative samples are needed to discover the similarity and dissimilarity, which requires large GPU memories \cite{Chen2020Simple_arxiv}. To address this problem, SimCLR \cite{Chen2020Simple_arxiv} employs multi-machine distributed computing to enlarge the batch. Nevertheless, due to GPU memory limitation, further increasing positive/negative samples is prohibitive in practice, which forms a barrier to improving SSL. We identify this as an under-clustering problem.

To avoid under-clustering, more elegantly, Mean Teacher \cite{Tarvainen2017Mean_nips} is applied to produce sufficient negative \cite{He2020Momentum_cvpr,Chen2020Improved_arxiv} and positive samples \cite{Grill2020Bootstrap_arxiv,Caron2020Unsupervised_arxiv}. Moreover, Exemplar-CNN \cite{Dosovitskiy2014Discriminativ_neurips,Dosovitskiy2016Discriminative_pami} directly clarifies each image in a dataset into a cluster, i.e., it categorizes CIFAR-10 into 50K clusters. However, it obtains an unsatisfactory performance. We identify this as an over-clustering problem. Specifically, since each image can be regarded as a cluster, excessive negative sample pairs can over-cluster samples from the same category into different clusters. This over-clustering can lead to bad representation learning since the network just memorizes the data instead of learning from the data \cite{Zhang2017Understanding_iclr,bresler2020corrective}.

To reduce over-clustering, recent works rethink the necessity of negative samples and propose to remove negative samples at all. Notable works include BYOL \cite{Grill2020Bootstrap_arxiv} and SimSiam \cite{Chen2021Exploring_cvpr}. However, once the negative samples are removed, under-clustering \emph{may probably} reoccur because the model cannot efficiently discover the dissimilarity between inter-class samples (see Table \ref{tab:intro} for the evidence). Besides, a few other pioneering works are also proposed to analyze the negative samples' role in the contrastive loss. \cite{Cai2020Are_arxiv} used empirical evidence to show that not all negatives are equally important for contrastive learning. \cite{Huynh2020Boosting_arxiv} used a complicated way to cancel false negatives. \cite{Chuang2020Debiased_neurips,Robinson2021Contrastive_iclr} observed that using extremely close samples is bad for contrastive learning and leverage distribution knowledge to solve the problem. As opposed to these methods that use over-complicated contrastive losses, we use a quite simple triplet loss.

\textbf{Triplet loss.} Triplet loss was proposed by Ding et al. \cite{Ding2015Deep_pr} and  Schroff et al. \cite{Schroff2015FaceNet_cvpr} independently for person re-identification and face recognition, respectively. It tends to maximize the relative distance between the positive pair and the negative pair for each triplet unit. Several improvements over triplet loss are conducted to discover the valuable triplets \cite{Hermans2017Defense_arxiv}, to perform cross-batch triplet loss \cite{Wang2020Cross_cvpr}, and to apply to weakly supervised scenario \cite{Wang2020Weakly_tnnls}. However, these classical triplet losses can also result in over-clustering. In contrast, we propose a truncated triplet loss to address the over-clustering problem guaranteed by the Bernoulli Distribution model.

Recognizing the hardest negatives can lead to bad local minima in practice, \cite{Schroff2015FaceNet_cvpr} proposes a semi-hard negative sampling strategy, sharing the merit of our method in avoiding over-trusting the hardest negative sample and benefiting representation learning. The differences between our truncated triplet loss and semi-hard sampling strategy are two-fold. First, if we read the widely-used code in TensorFlow and Pytorch, we can find that a semi-hard triplet loss is a margin-based loss rather than a ranking loss. Second, our method trims negative samples while \cite{Schroff2015FaceNet_cvpr} performs vanilla sampling. Our approach differs from \cite{Schroff2015FaceNet_cvpr}'s vanilla sampling and mixture sampling and is thus a new one.

\section{Self-supervised Representation Learning}

We first present the under-clustering and over-clustering problems of contrastive learning in Section \ref{sec:problem}.
Then, we present our method in Section \ref{sec:truncated}. The analysis of the effectiveness of our approach is presented in Section \ref{sec:analysis}.

\subsection{Under-clustering and Over-clustering}\label{sec:problem}

Contrastive learning is proposed by \cite{Hadsell2006Dimensionality_cvpr} and is widely used in SSL, achieving best-performing results on ImageNet. The most widely-adopted loss for contrastive learning is InfoNCE \cite{Oord2018Representation_arxiv}. Let $x$ be a query image which has 1 positive sample $x^+$ and $m$ negative samples $\{x_j^-\}_{j=1,\cdots, m}$. InfoNCE calculates their inner products and normalize the products using softmax and have:\begin{small}\begin{equation}
\begin{aligned}
&\{\widetilde{x^T x^+}, \widetilde{x^T x_1^-}, \widetilde{x^T x_2^-}, \cdots, \widetilde{x^Tx_m^-}\}\\
=&softmax (\{x^T x^+, x^T x_1^-, x^T x_2^-, \cdots, x^Tx_m^-\}).
\end{aligned}\nonumber
\end{equation}\end{small}Then, the goal of contrastive learning is to minimize:\begin{small}\begin{equation}
- 1 \log \widetilde{x^T x^+} - 0 \log\widetilde{x^T x_1^-} - 0\log\widetilde{x^T x_2^-}\cdots- 0\log\widetilde{x^Tx_m^-}.\nonumber
\end{equation}
\end{small}(see footnote\footnote{Generally, it's written as: \begin{scriptsize}$-\log \frac{\exp(x^T x^+)/\tau}{\exp(x^T x^+)/\tau + \sum\limits_{j=1}^{m}\exp(x^T x_j^-)/\tau}$\end{scriptsize}, where $\tau$ is a temperature.}), which can be interpreted as forcing \begin{small}$\widetilde{x^T x^+}$\end{small} to be close to 1 and forcing \begin{small}$\widetilde{x^T x_1^-} $, $\widetilde{x^T x_2^-}$, ..., $\widetilde{x^Tx_m^-}$\end{small} to be close to 0. This indicates that plenty of negative sample pairs are needed to guarantee the learning efficiency because the model needs sufficient negative samples to discover the dissimilarity between inter-class samples. Especially, plenty of positive samples and negative samples are needed to enrich the similarity and dissimilarity \emph{in each batch of data}.
 
\textbf{Under-clustering.} Insufficient positive and negative examples can lead to under-clustering. Under-clustering is a critical problem in which different categories have a valid (but unwelcome) overlap. For example, in Figure \ref{fig:problem} (a), a cluster may contains \{dog, horse\} or \{cat, cow\}, i.e., dogs and horse are mixed up. Without the annotation, we cannot identify the actual label of each data point. In other words, the dogs and horses have an overlap. An under-clustering problem occurs when insufficient positive and negative samples are present.

\textbf{Over-clustering.} As opposed to under-clustering caused by lacking negative samples, over-clustering is caused by overwhelming negative samples. Although contrastive learning implicitly regards each image as a class, we do not expect over-clustering. Excessive negative sample pairs can result in over-clustering that forces samples from the same category into different clusters. As is shown in Figure \ref{fig:problem} (c), if excessive negative examples are provided, the two dogs that belong to the same category are now assigned to two clusters. Similar phenomena also appear in cats and cows. This non-ideal over-clustering would prevent the model from learning discriminative representations summarizing essential features of a category since the network just memorizes the data instead of learning from the data \cite{Zhang2017Understanding_iclr,bresler2020corrective}.

\begin{figure*}
\centering
\includegraphics[width=0.75\textwidth]{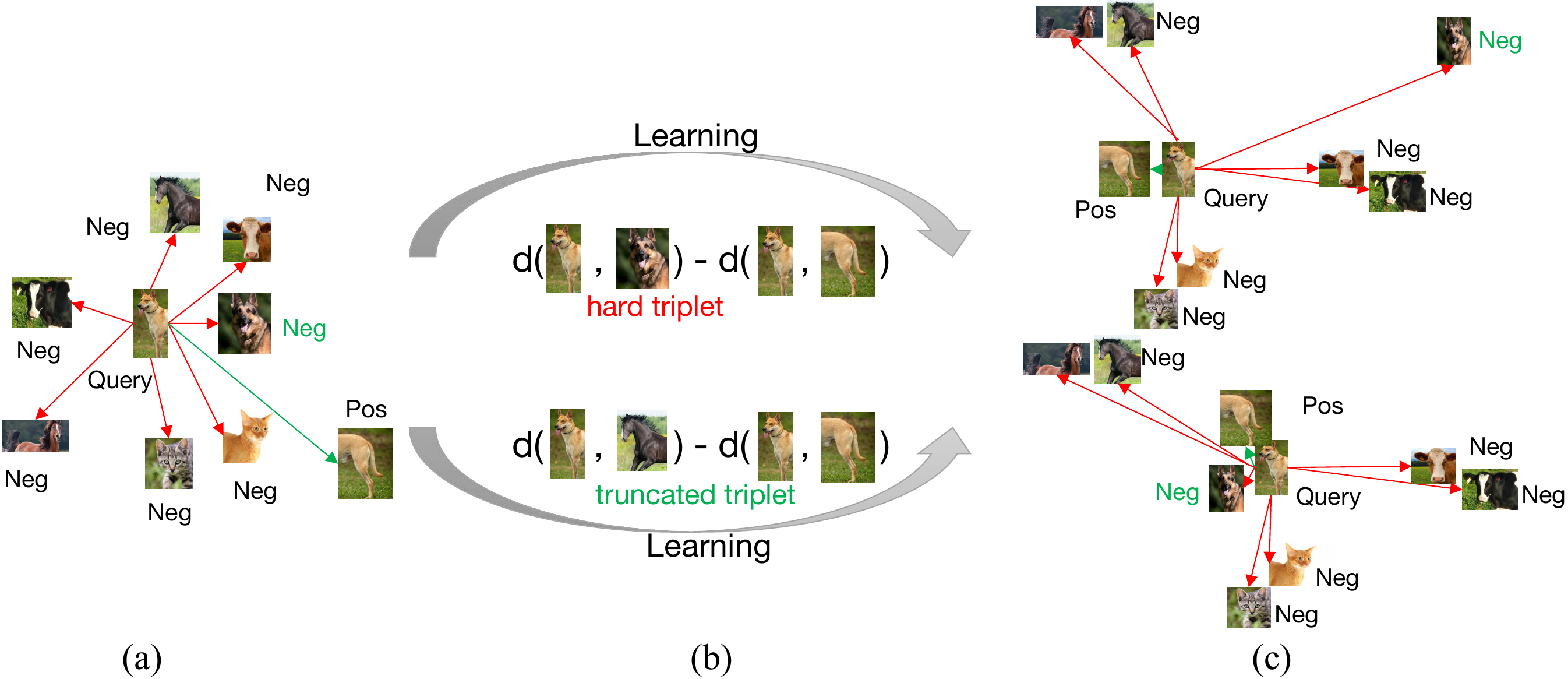}
\vspace{-3pt}
\caption{Illustration of our truncated triplet loss. Each sample pair connected by a red line represents a negative sample pair. Note that although labeled by a green word ``Neg'', in fact, the dog is a positive sample to the query image because they belong to the same category, i.e., ``a dog''. Traditional triplet loss using the hardest triplet (see Figure (b)) results in over-clustering, in which the distance between these two dogs will be enlarged. The over-clustering result is shown at (c) \textbf{top}, and the ideal learning results is shown at (c) \textbf{bottom}.}\label{fig:triplet}
\vspace{-15pt}
\end{figure*}

\textbf{Ideally}, we would like to use just the right amount of negative sample pairs to ensure that the images from the same category are close to each other and that the images from different classes are far away. As shown in  Figure \ref{fig:problem} (b), all the dogs, cats, and cows are clustered correctly. Note that this is achieved in an unsupervised manner.

\subsection{Truncated Triplet Loss}\label{sec:truncated}

\textbf{Triplet loss.} Inspired by relative distance comparison \cite{Zheng2013Reidentification_pami}, triplet loss was proposed by \cite{Ding2015Deep_pr} and \cite{Schroff2015FaceNet_cvpr} independently for person re-identification and face recognition, respectively. In a triplet-loss method, a set of triplets, i.e., $\{(x_i, x_i^+, x_i^-)\}_{i=1,\cdots, m}$, are first generated. In general, a query image will have far more negative samples than positive samples (see Figure \ref{fig:triplet} (a) for detail). For presentation simplicity, we use only one query image and one positive sample for illustration, i.e., we have a triplet set $\{(x, x^+, x_i^-)\}_{i=1,\cdots, m}$. The oldest triplet loss is defined as:\begin{small}\begin{equation}
\mathcal{L}oss = {\sum}_{i=1}^{m}\max\big(d(x, x^+) - d(x, x_i^-), \mathcal{C}\big)\nonumber,
\end{equation}\end{small}where $d$ is a distance metric (e.g., cosine distance or Euclidean distance). Here, $\mathcal{C}$ is a margin deciding whether or not to drop a triplet. This is critical in machine learning algorithms since we usually drop the simple data and focus on the hard data near the decision boundary, as support vector machine \cite{cortes1995support} suggests. To improve the learning efficiency of triplet loss, in practice, we usually use the hardest triplet to represent the overall triplets, i.e., only the triplet containing the negative sample of the highest similarity score overall negative samples are used (please refer to Figure \ref{fig:triplet} (b) for detail). Finally, a triplet is formally defined as:\begin{small}\begin{equation}
    \mathcal{L}oss = \max\big(d(x, x^+) - d(x, x_{hardest}^-), \mathcal{C}\big).
\end{equation}\end{small}Since $x_{hardest}^-$ is the hardest negative sample, we have $d(x, x_{hardest}^-) \le d(x, x_{i}^-)$ for all $i$. This indicates that when the hardest triplet loss meets the condition $d(x, x^+) \le d(x, x_{hardest}^-)$, all others triplets meet the condition. Therefore, the hardest triplet loss guarantees a considerable distance between negative sample pairs. Using triplet loss, we can address the under-clustering problem.

Although triplet loss largely reduces under-clustering, it increases over-clustering risk. Specifically, since contrastive learning can be considered a classification problem that identifies each image as a class, using the hardest triplet loss can lead to over-clustering. For example, in Figure \ref{fig:triplet} (a), the two dogs belong to the same object category. Unsurprisingly, their feature similarity is high. But in SSL, the actual category labels are absent; thus, these two dogs can be reluctantly considered negative sample pairs (Figure \ref{fig:triplet} (b) and (c), \textbf{top}). This indicates they are the hardest negative sample pairs. Using the hardest triplet loss, the distance between these two dogs is enlarged. This results in an over-clustering problem in that the two dogs from the same category are over-clustered into two different clusters.

\textbf{Truncated triplet loss.} To avoid over-clustering, we construct the negative pair by truncating/trimming the hardest negative samples. We select a negative sample deputy to form a truncated triplet, i.e., we have:\begin{equation}
    \mathcal{L}oss = \max\big(\gamma d(x, x^+) - d(x, x_{deputy}^-), \mathcal{C}\big).
\end{equation}Specifically, $d(x, x_{deputy}^-)$ is obtained using the following steps. First, we compute the distance $\{d(x, x_i^-)\}, \forall i$. Then, we sort $\{d(x, x_i^-)\}$ by ascending. Finally, we obtain $d(x, x_{deputy}^-)$ in two ways:\begin{itemize}
\item \emph{rank-k triplet loss:} the $k$-th element are selected from $\{d(x, x_i^-)\}$, yielding: $d(x, x_{deputy}^-) = d(x, x_{rank-k}^-)$. 
\item \emph{smoothed-rank-k triplet loss:} selecting the top-2, top-3, ..., top-(2k+1) elements from $\{d(x, x_i^-)\}$ and yielding: \begin{small}$d(x, x_{deputy}^-)= \frac{1}{2k}\sum\limits_{j=2}^{2k+1} d(x, x_{rank-j}^-)$\end{small}.
\end{itemize}Note that when $k=1$, the rank-k triplet loss reduces to the hardest triplet loss. We can show in Section \ref{sec:analysis} that replacing the hardest triplet with the rank-k triplet can indeed reduce the risk of over-clustering, guaranteed by the Bernoulli Distribution model. By default, we use $k=\frac{m}{2}$ (i.e., the triplet deputy) for most of the experiments, although using other values (e.g., $k=5$) also yields good performance. We use the widely-used cosine distance for $d$, i.e., $d(x,y) = - \frac{xy}{\|x\|_2 \|y\|_2}$, where $\|\cdot\|_2$ represents the L2 norm. The negative sign (``-'') is used here because we usually consider that a higher similarity indicates a smaller distance. We set $\gamma$ to be 2. \emph{Note that, according to our experiments, the smoothed-rank-k triplet loss achieves better performance than the rank-k triplet loss.}

\subsection{Analysis with Bernoulli Distribution Model }\label{sec:analysis}

In contrastive learning, views from different images are considered negative sample pairs even they are from the same actual category (e.g., the two dogs in Figure \ref{fig:triplet} (a)). Unsurprisingly, these kinds of negative sample pairs have a high feature similarity. With a high probability, they will be in the hardest triplets (see Figure \ref{fig:triplet} (b)). Using the hardest triplet loss, the distance between these false-negative pairs is enlarged. This results in an over-clustering problem that the false-negative pairs from the same category are over-clustered into two different clusters (see Figure \ref{fig:triplet} (c) \textbf{top}).

But in our truncated triplet loss, we sort $\{d(x, x_i^-)\}$ by ascending and select the $k$-th element from $\{d(x, x_i^-)\}$ to form $d(x, x_{deputy}^-)$. An over-clustering risk exists if this rank-k negative sample and the query image belong to the same actual category. We need to estimate the probability that these two images belong to the same category. We first have a reasonable assumption: with a high probability, the image pairs from the same category have higher feature similarities than other pairs, and the distances between these pairs are smaller than other pairs. Because we have sorted $\{d(x, x_i^-)\}$ by ascending, the event that the rank-k negative sample and the query belong to the same category indicates an event that at least $k$ negative samples and the query belong to the same category. The probability of this event can be computed by using a Bernoulli Distribution model, i.e., \begin{small}$\Pr = \sum\limits_{j=k}^{m}\mathbf{C}_m^j p^j (1-p)^{m-j}$ \end{small}. Here, $\mathbf{C}$ is used to denote the combinations, and $p$ is used to denote the probability that a negative sample and the query belong to the same class. For example, on ImageNet, we have $p=\frac{1}{1000}$. In our experiment, we let $m$ be 104 and $k$ be $\frac{m}{2}$. Putting $m,k$ into the above equation, we have $\Pr = 6.53e^{-121}$, which is almost zero. Even we let $m$ be 104 and $k$ be 5, we have $\Pr = 3.03e^{-94}$. This indicates it is a rare event that the rank-k negative sample and the query belong to the same category. Thus, our truncated triplet loss can avoid over-clustering, guaranteed by the Bernoulli Distribution model, e.g., the two dogs at the bottom of Figure \ref{fig:triplet} (c) can be identified correctly. Experimental results in Section \ref{sec:ablation} also verify the effectiveness of our method\footnote{The Bernoulli Distribution model is simple yet can be justified by the experiment (Section \ref{sec:ablation}). Meanwhile, \cite{Chuang2020Debiased_neurips,Robinson2021Contrastive_iclr} used elegant mathematical boundedness tools to show the generalization ability of contrastive learning. A more elegant justification using boundedness tools is welcome.}.

\section{Main results}\label{sec:main}

Our SSL training protocols are as follows.

\textbf{Data augmentation protocol.} Our augmentations are straightforward, including randomly cropping, randomly resizing, randomly flipping horizontally, arbitrary gray scaling, stochastic color jittering, Gaussian blurring, and solarization. Please see our codes.

\textbf{Other protocols.} In the unsupervised learning stage, the batch size is 104 images per GPU, and we use eight GPUs. The gradient update interval is five steps. The maximum epoch is 200. The learning rate starts from 4.8 and gradually decreases with cosine annealing. The weight decay factor is $1e^{-6}$. The optimizer is LARS \cite{ginsburg2018large} with a momentum 0.9. The backbone is ResNet-50, which is the same as the previous methods. The models are trained by using the 1.28M training images of ImageNet but without their annotations. The protocols are in line with \cite{OpenSelfSup2020}: we use the same momentum network, the same multilayer-perceptron head \& neck as BYOL. Also, following BYOL, our loss is symmetric \emph{w.r.t} the positive pairs. Using protocols similar to \cite{OpenSelfSup2020} makes it feasible to present comparisons on multiple datasets/tasks without the extra hyper-parameter search.

We evaluate our method by comparing it with state-of-the-art methods in four tasks, involving linear evaluation on ImageNet, person re-identification on SYSU-30k, and object detection on both COCO 2017 and ``VOC07+12''.

\subsection{Linear evaluation on ImageNet} 

Linear evaluation is the most widely adopted evaluation protocol for validating the representation ability of different SSL methods. Standardly, the backbones of ResNet-50 are trained by using the above SSL training protocols and are frozen. A linear classifier is then added to the top of the frozen representation and is trained for each method. All methods are trained using the 1.28M training images of ImageNet and are evaluated using the 50K validation images of ImageNet. For the linear classification stage, the batch size is 256. The maximum epoch is 100. There is no weight decay in linear classification training. The optimizer is SGD. Single-scale center-crop top-1 accuracy is used.

\begin{table}
\centering
\scriptsize
\caption{Top-1 accuracy and training epochs of state-of-the-art methods on ImageNet using linear classification for evaluation.}\label{tab:imagenet}
\begin{tabular}{l|ccc}
Method & top-1 acc. & train epochs\\
\hline
Random & 4.4 & 0\\
Relative-Loc \cite{Doersch2015Unsupervised_iccv} & 38.8 & 200\\
Rotation-Pred \cite{Gidaris2018Unsupervised_iclr}& 47.0 & 200\\
DeepCluster \cite{Caron2018Deep_eccv} & 46.9 & 200\\
NPID	 \cite{Wu2018Unsupervised_arxiv} & 56.6 & 200 \\
ODC \cite{Zhan2020Online_cvpr} & 53.4 & 200\\
SimCLR  \cite{Chen2020Simple_arxiv}  & 60.6 & 200\\
SimCLR \cite{Chen2020Simple_arxiv}  & 69.3 & 1000 \\
MoCo \cite{He2020Momentum_cvpr} & 61.9 & 200\\
MoCo v2 \cite{Chen2020Improved_arxiv} & 67.0  & 200\\
MoCo v2 \cite{Chen2020Improved_arxiv} &  71.1 & 800 \\
SwAV \cite{Caron2020Unsupervised_arxiv} (single-crop)   & 69.1  & 200\\
SwAV \cite{Caron2020Unsupervised_arxiv} (multi-crop)    & 72.7  & 200\\
BYOL \cite{Grill2020Bootstrap_arxiv} & 71.5  & 200\\
BYOL \cite{Grill2020Bootstrap_arxiv} &  72.5  & 300 \\
BYOL \cite{Grill2020Bootstrap_arxiv} &  74.3  & 1000 \\
SimSiam \cite{Chen2021Exploring_cvpr} & 68.1 & 100 \\ 
SimSiam \cite{Chen2021Exploring_cvpr} & 70.0 & 200 \\
SimSiam \cite{Chen2021Exploring_cvpr} & 70.8 & 400 \\
SimSiam \cite{Chen2021Exploring_cvpr} & 71.3 & 800 \\
\hline
\textbf{\textcolor{Mycolor1}{truncated triplet}} & \textbf{\textcolor{Mycolor1}{73.6}} &  \textbf{\textcolor{Mycolor1}{180}} \\ 
\textbf{\textcolor{Mycolor1}{truncated triplet (smoothed)}} & \textbf{\textcolor{Mycolor1}{73.8}} &  \textbf{\textcolor{Mycolor1}{200}} \\ 
\textbf{\textcolor{Mycolor1}{truncated triplet (multi-crop)}} & \textbf{\textcolor{Mycolor1}{74.1}} &  \textbf{\textcolor{Mycolor1}{200}} \\ 
\textbf{\textcolor{Mycolor1}{truncated triplet}} & \textbf{\textcolor{Mycolor1}{75.9}}  &  \textbf{\textcolor{Mycolor1}{700}} \\ 
\hline
supervised & 76.3 & 100\\
supervised + linear eval & 74.1 & 100\\
supervised & 78.4 & 270\\
\end{tabular}
\vspace{-20pt}
\end{table}

Currently, the widely-used evaluation standard of SSL merely values the accuracy but regardless of the training epochs. Following this standard, we first compare our method with the competitors without considering the training epochs. Table \ref{tab:imagenet} shows that our method achieves a promising result on ImageNet, i.e., 75.9\%, outperforming the latest state-of-the-art methods by a clear margin.

Regarding learning efficiency, existing SSL methods' efficiency is lower than supervised learning. As shown in Table \ref{tab:imagenet}, SSL models are trained for about 1,000 epochs, while the supervised counterpart is merely trained for 100 epochs. SimCLR \cite{Chen2020Simple_arxiv} explains that training for a longer time does not bring gain for the supervised learning model (i.e., it reported a result of 76.4\% vs. 76.3\% for 1000 epochs' vs. 100 epochs' supervised training). But our observation is the opposite, i.e., our reproduction of a supervised model trained for 270 epochs can achieve 78.4\% top-1 accuracy, which is significantly higher than all of the SSL models.

Put together, a complete comparison regarding both accuracies and training epochs is presented in Figure \ref{fig:intro} and Table \ref{tab:imagenet}, where we have three observations. \textbf{First}, previous SSL methods still have a long way to go. They have significantly lower learning efficiency than supervised learning. \textbf{Second}, as shown in Figure \ref{fig:intro}, our approach lies in the top-left corner of the figure, indicating our method achieves the best performance among the compared SSL methods. For example, SwAV \cite{Caron2020Unsupervised_arxiv} achieves 72.7\% (200 epochs), which is lower than our method (73.6\%, 180 epochs). \textbf{Note} that SwAV uses an additional multi-crop augmentation that we don't use, which has a 3.6\% gain\footnote{Many methods (e.g., HSA \cite{Xu2020Hierarchical_arxiv} / MoCo v2 \cite{Chen2020Improved_arxiv} / SimCLR \cite{Chen2020Simple_arxiv} / SeLa \cite{Asano2020Self_iclr} / DeepCluster \cite{Caron2018Deep_eccv}) can benefit significantly from multi-crop augmentation, but BYOL \cite{Grill2020Bootstrap_arxiv} cant't.} (without multi-crop augmentation, SwAV only achieves 69.1\%). For a fair comparison, we add multi-crop augmentation to our method, leading to a further state-of-the-art result of 74.1\% (200 epochs). This comparison verifies the effectiveness and efficiency of our approach. \textbf{Third}, the smoothed truncated triplet loss achieves better performance than the truncated triplet loss.

\begin{table}
\centering
\scriptsize
\caption{Object detection results on COCO 2017 for Mask-RCNN.}\label{tab:coco}
\begin{tabular}{l|c|c}
Method & AP$^{Box}$ & AP$^{Mask}$  \\
\hline
Random & 35.6 & 31.4\\
Relative-Loc \cite{Doersch2015Unsupervised_iccv} &  40.0 & 35.0 \\
Rotation-Pred \cite{Gidaris2018Unsupervised_iclr}& 40.0 &34.9 \\
NPID	 \cite{Wu2018Unsupervised_arxiv} & 39.4 & 34.5 \\
MoCo \cite{He2020Momentum_cvpr} & 40.9 & 35.5  \\
MoCo v2 \cite{Chen2020Improved_arxiv}  & 40.9 & 35.5 \\
SimCLR  \cite{Chen2020Simple_arxiv}  & 39.6 & 34.6 \\
BYOL \cite{Grill2020Bootstrap_arxiv} & 40.3 & 35.1\\
\hline
\textbf{\textcolor{Mycolor1}{truncated triplet}}  & \textbf{\textcolor{Mycolor1}{41.7}} & \textbf{\textcolor{Mycolor1}{36.2}}  \\ \hline
supervised & 40.0  & 34.7 \\
\end{tabular}
\vspace{-11pt}
\end{table}

\begin{table}
\centering
\scriptsize
\caption{Object detection results on VOC07+12 for Faster-RCNN.}\label{tab:voc}
\begin{tabular}{l|ccc}
Method & AP50$^{Box}$& AP$^{Box}$ & AP75$^{Box}$  \\
\hline
Random & 59.0 & 32.8 & 31.6\\
Relative-Loc \cite{Doersch2015Unsupervised_iccv} & 80.4 & 55.1 & 61.2 \\
Rotation-Pred \cite{Gidaris2018Unsupervised_iclr}& 80.9 & 55.5 & 61.4 \\
NPID	 \cite{Wu2018Unsupervised_arxiv} & 80.0 & 54.1 & 59.5 \\
MoCo \cite{He2020Momentum_cvpr} & 81.4 & 56.0 & 62.2 \\
MoCo v2 \cite{Chen2020Improved_arxiv}  & 82.0 & 56.6 & 62.9\\
SimCLR  \cite{Chen2020Simple_arxiv}  & 79.4 & 51.5 & 55.6\\
BYOL \cite{Grill2020Bootstrap_arxiv} & 81.0 & 51.9 & 56.5 \\\hline
\textbf{\textcolor{Mycolor1}{truncated triplet}} & \textbf{\textcolor{Mycolor1}{82.6}} &  \textbf{\textcolor{Mycolor1}{56.9}} & \textbf{\textcolor{Mycolor1}{63.8}} \\ \hline
supervised &81.6 & 54.2 & 59.8 \\
\end{tabular}
\vspace{-15pt}
\end{table}

\subsection{Transferring to downstream tasks}

\textbf{Transferring to COCO 2017 object detection.} One of SSL's goals is to learn transferrable features. We test our learned representation's generalization ability by transferring to COCO 2017 object detection \cite{Lin2014Microsoft_eccv}, \emph{which is (one of) the largest benchmarks for general object detection,} containing about 119K training images. Specifically, the backbones of ResNet-50 are trained by using the above SSL training protocols, and the trained network weights serve as the initialization of Mask-RCNN \cite{He2017Mask_iccv} with C4. We fine-tune all layers on the \emph{train2017} set. The training schedule is the default 2$\times$ schedule in \cite{Detectron2018}. As suggested by \cite{He2020Momentum_cvpr}, the distribution of features obtained by unsupervised pre-training and supervised pre-training is not the same. Hence, we set a higher learning rate than supervised counterparts for finetuning. We also finetune BN instead of freezing it following \cite{He2020Momentum_cvpr}. The accuracy is tested on the \emph{val2017} set. We report the standard metric for the detection and instance segmentation: AP$^{\text{Box}}$ and AP$^\text{Mask}$.

Table \ref{tab:coco} shows that using our approach for pretraining surpasses other SSL ImageNet pretraining on COCO 2017 detection. \emph{Our method indeed surpasses prior arts (including MoCo / MoCo v2) on object detection on COCO 2017.} Moreover, our SSL pretraining even outperforms the supervised ImageNet pretraining, implying that SSL can obtain more universal representations. This is in line with previous works that also show that SSL pretraining can outperform supervised pretraining on object detection \cite{Gidaris2020Learning_cvpr,He2020Momentum_cvpr,Misra2020Self_cvpr,Caron2020Unsupervised_arxiv}.

\textbf{Transferring to VOC07+12 object detection.} In addition to COCO 2017, we also evaluate our method's transferability in PASCAL VOC object detection. Following \cite{He2020Momentum_cvpr} and \cite{OpenSelfSup2020}, the backbones of ResNet-50 are trained by using the above SSL training protocols, and the trained network weights serve as the initialization of Faster R-CNN \cite{Ren2015Faster_neurips} with C4. Then, we fine-tune all layers on the \emph{trainval07+12} set of PASCAL. The image scale is [480, 880] pixels during training and 800 in the testing. We report the default VOC metric of AP$^\text{50}$ and the COCO-style AP and AP$^\text{75}$. The evaluation is on the VOC \emph{test2017} set.

We show the results of different methods in Table \ref{tab:voc}. As shown, our approach achieves state-of-the-art performance. Note that only MoCo v2 and our approach can catch up with the supervised pretraining counterpart that performs pretraining for 100 epochs. This verifies the effectiveness of our method and implies that our SSL can obtain universal representations.

\subsection{Person re-identification on SYSU-30k}

In a general sense, all the above tasks (image classification, object detection, and segmentation) belong to visual categorization since detection and segmentation can be considered categorizing regions and pixels for a given image. The effectiveness of our approach beyond visual classification remains uninvestigated. In the following, we investigate a different task, i.e., person re-identification (re-ID), which is fundamental in video surveillance \cite{farenzena2010person}. Re-ID refers to the problem of re-identifying individuals across cameras \cite{Wang2020Grammatically_kdd}. Mathematically, re-ID is a matching problem rather than a classification problem because it requires calculating distance metrics between two given images. As is proved by \cite{fan2018unsupervised,wang2020smoothing,Wang2020Weakly_tnnls,Zhuo2018Occluded_icme,Wang2019Spatial_aaai}, unsupervised representation learning is critical to visual matching; therefore, validating the effectiveness of our approach in re-ID is nontrivial.

\textbf{Dataset and protocol.} We conduct experiments on the SYSU-30k dataset \cite{Wang2020Weakly_tnnls}, which is the largest database for re-ID. This database contains 29,606,918 images of 30,508 pedestrians, which is about 30 times larger than ImageNet in terms of category number. Please note that the exact label of each image is unknown in this dataset. Both the lack of precise annotation and the massive number of images make this dataset set very suitable for unsupervised learning, especially SSL. Since we are the first to perform SSL on this database, no previous work provides an evaluation protocol for this dataset. We propose a new evaluation protocol for it: the training set of SYSU-30k is employed to perform SSL. Once the model is learned, we directly use it to extract features for visual matching on the test set of SYSU-30k without any finetuning. This is even more challenging than linear evaluation on ImageNet because linear evaluation learns an extra classifier for recognition, but no extra classifier is learned here. Regarding the superiority of using SYSU-30k mentioned above, we believe SYSU-30k is a perfect database to evaluate SSL effectiveness and recommend it to future SSL researchers.

\begin{table}
\centering
\small
\caption{Comparison with state-of-the-art methods on SYSU-30k.}\label{tab:sysu-30k}
\begin{tabular}{l|cc}
supervision  & method & rank-1  \\
\hline
\multirow{4}{*}{Transfer learning} & DARI \cite{Wang2016DARI_aaai} & 11.2\\
 & DF \cite{Ding2015Deep_pr} & 10.3\\
& Local CNN \cite{Yang2018Local_acmmm} & 23.0\\
& MGN \cite{Boll2018Learning_acmmm} & 23.6\\\hline
\multirow{2}{*}{Weakly supervised} & W-Local CNN \cite{Wang2020Weakly_tnnls} & 28.8\\
&W-MGN \cite{Wang2020Weakly_tnnls} & 29.5\\\hline
\multirow{4}{*}{Self-supervised} & SimCLR \cite{Chen2020Simple_arxiv} & 10.9\\
& MoCo v2 \cite{Chen2020Improved_arxiv} & 11.6 \\
 & BYOL \cite{Grill2020Bootstrap_arxiv} & 12.7 \\
 & \textbf{\textcolor{Mycolor1}{truncated triplet}} & \textbf{\textcolor{Mycolor1}{14.8}} \\
\end{tabular}
\vspace{-22pt}
\end{table}

\textbf{Result analysis.} We compare our method with SimCLR, MoCo-v2, BYOL, and current state-of-the-art results (not SSL methods). We use ResNet-50 as the backbone. The results in Table \ref{tab:sysu-30k} show that our models achieve new state-of-the-art performance, i.e., a rank-1 accuracy of 14.8\%. Note that this number is low, i.e., even lower than existing transfer learning and weakly supervised learning methods. This is attributed to the challenge of the SYSU-30k test set, which contains about 480,000 \textbf{testing} images. Moreover, there are 478,730 mismatching images as the wrong answer in the gallery. Thus, evaluation using the SYSU-30k test set is like searching for a needle in a haystack. We encourage future SSL researchers to use this dataset to evaluate the effectiveness of SSL. We can also observe that our approach surpasses other SSL methods by a clear margin (14.8 vs. 12.7 for ours vs. BYOL). This verifies the effectiveness of our approach on visual matching tasks like re-ID.

\section{Ablation studies}\label{sec:ablation}

\begin{table}
\small
\centering
\caption{Effect of avoiding over-clustering.}\label{tab:over-clustering}
\begin{tabular}{l|c|cc}
Training epoch & event & 0  &  180  \\
\hline
\multirow{2}{*}{$k=1$} & $\Pr(\Omega| \mathcal{A})$   & 0.1538 &    0.9656   \\
 & $\Pr(\Omega| \mathcal{B})$  & 0.1618   &  0.9948   \\\hline
\multirow{2}{*}{\textcolor{Mycolor1}{$k=5$} } & \textcolor{Mycolor1}{$\Pr(\Omega| \mathcal{A})$}  & \textcolor{Mycolor1}{0.1167}  &  \textcolor{Mycolor1}{0.2105} \\
& \textcolor{Mycolor1}{$\Pr(\Omega| \mathcal{B})$}  & \textcolor{Mycolor1}{0.1230}    & \textcolor{Mycolor1}{0.2132} \\\hline
\multirow{2}{*}{\textcolor{Mycolor1}{$k=52$}} & \textcolor{Mycolor1}{$\Pr(\Omega| \mathcal{A})$}  &  \textcolor{Mycolor1}{0.1220}   & \textcolor{Mycolor1}{0.0110} \\
& \textcolor{Mycolor1}{$\Pr(\Omega| \mathcal{B})$}  &  \textcolor{Mycolor1}{0.1288}  & \textcolor{Mycolor1}{0.0233} \\
\end{tabular}
\vspace{-11pt}
\end{table}

\begin{table}
\small
\centering
\caption{Impact of margin.}\label{tab:margin}
\begin{tabular}{l|ccc}
Margin & $\mathcal{C}=-0.3$ & \textcolor{Mycolor1}{$\mathcal{C}=-1.2$} &  \textcolor{Mycolor1}{$\mathcal{C}=-100$}  \\
\hline
Top-1 accuracy & 28.3 & \textcolor{Mycolor1}{29.8}  & \textcolor{Mycolor1}{30.0} \\
\end{tabular}
\vspace{-7pt}
\end{table}

\begin{table}
\small
\centering
\caption{Impact of rank-k.}\label{tab:rank-k}
\begin{tabular}{l|ccc}
Rank-k  & rank-1 & \textcolor{Mycolor1}{rank-5} &  \textcolor{Mycolor1}{rank-52}  \\
\hline
Top-1 accuracy & 28.9  & \textcolor{Mycolor1}{29.5}  & \textcolor{Mycolor1}{30.0}  \\
\end{tabular}
\vspace{-16pt}
\end{table}

In the rest of our paper, to reduce the training time and fast access to the results, we perform ablation studies using 20 training epochs. Please note that, due to our method's high learning efficiency, training for 20 epochs is sufficient for ablation studies. The linear evaluation training is also reduced to one epoch \footnote{After running plenty of experiments, our practical experience shows that 20 epochs of training are enough for ablation studies.}. Actually, previous works also use few training epochs for ablation studies, e.g., \cite{Chen2020Simple_arxiv} and \cite{Tian2020What_arxiv}. All the training protocols are the same as Section \ref{sec:main}, except that we take the 20th epoch's checkpoint for one epoch's evaluation. This section only reports the results of top-1 accuracy on ImageNet under the linear evaluation protocol since it is the most widely adopted metric for validating the effectiveness of SSL methods.

\textbf{Effect of avoiding over-clustering.} As we discussed in Section \ref{sec:truncated}, thanks to the truncated triplet loss, we can avoid over-clustering. For example, if $k=5$, the probability of over-clustering is $3.03e^{-94}$. If $k=\frac{m}{2}=52$, the probability of over-clustering is $6.53e^{-121}$. However, whether this analysis is correct remains unclear. In the following, we provide empirical analysis. During batch training, all the batch samplings are considered the total event $\mathcal{A}$. If a batch contains at least two images belonging to the same actual category, we call it an event $\mathcal{B}$. If (at least) these two images are mis-considered a false-negative pair, we call this an event $\Omega$. We report the frequency $\Pr(\Omega|\mathcal{A})$ and $\Pr(\Omega|\mathcal{B})$ in Table \ref {tab:over-clustering} for different training epochs and different $k$s.

We have three observations from Table \ref{tab:over-clustering}. \textbf{First},  the rank-52 and rank-5 negative samples rarely belong to the same category as the query image, i.e., $\Pr(\Omega|\mathcal{A})$ and $\Pr(\Omega|\mathcal{B})$  are low. \textbf{Second}, with the training epochs increasing, the probability that the rank-k negative sample is a false negative decreases. This is attributed to the more and more discriminative features that have been learned as the training goes. \textbf{Third}, with $k$ increasing, the probability that the rank-k negative sample is false negative decreases. Especially when $k=1$, our truncated triplet loss reduces to the hardest triplet loss. As shown, the hardest triplet loss indeed has a risk of over-clustering because the probability $\Pr(\Omega|\mathcal{B})$ is high. With $k$ increasing, the probability $\Pr(\Omega|\mathcal{B})$ decreases. This indicates that our truncated triplet loss can avoid the over-clustering problem, guaranteed by the Bernoulli Distribution model.

\emph{Please note that if a batch contains even one false negative sample pair, we consider the whole batch has the over-clustering risk. Therefore, the probability in Table \ref{tab:over-clustering} (e.g., 0.0110 or 0.2105) is higher than that in the analysis (e.g.,$3.03e^{-94}$ or $6.53e^{-121}$).}

\textbf{Impact of margin.} As we discussed in Section \ref{sec:truncated}, there is a margin $\mathcal{C}$ deciding whether or not to drop a triplet. This is critical in machine learning algorithms since we usually drop the simple data and focus on the complex data near the decision boundary, as support vector machine \cite{cortes1995support} suggests. To empirically verify this hypothesis, we train our method using different margins. The results are shown in Table \ref{tab:margin}. As shown, different margins lead to a performance fluctuation. When $\mathcal{C}=-100$ or $\mathcal{C}=-1.2$, the performance is best. Hence, we use $\mathcal{C}=-100$ for default in all of the experiments if no otherwise specified.

\textbf{Impact of rank-k.} As we analyze in Section \ref{sec:analysis}, we can use different $k$s for our truncated triplet loss. When $k=1$, our truncated triplet loss reduces to the traditional hardest triplet loss. When $k$ increases, the risks of over-clustering reduces exponentially.  To know the impact of using different $k$s, we train our method using different $k$s. As shown in Table \ref{tab:rank-k}, different $k$s lead to a performance fluctuation. When $k=5$ and $k=52$, the performances are satisfied. Hence, we use $k=5$ or $k=52$.  

\vspace{-2pt}
\section{Conclusion}
\vspace{-2pt}

Although SSL has shown promising results on ImageNet, its learning efficiency is still low. We attribute the inherited drawback of contrastive learning to under-clustering and over-clustering. To overcome these two problems, we propose a novel SSL framework using a truncated triplet loss. We employ triplet loss containing rich negative sample information to address the under-clustering problem. We trim negative samples to prevent the over-clustering problem, guaranteed by the Bernoulli Distribution model. Our method significantly improves the learning efficiency of SSL, leading to a state-of-the-art performance in several large-scale benchmarks and varieties of downstream tasks.

\vspace{-5pt}
\section*{Acknowledgement}
\vspace{-6pt}

This work is supported in part by the  National Key R\&D Program of China under Grant No. 2020AAA0109700, in part by the National Natural Science Foundation of China under Grant U1811463, 61836012, and 61876224, in part by the Natural Science Foundation of Guangdong Province of No. 2019A1515010939 and 2017A030312006, and in part by Innovate UK Smart Grant 33736.

{\small
\bibliographystyle{ieee_fullname}
\bibliography{egbib}
}

\end{document}